\documentclass{ecai}

\usepackage{latexsym}
\usepackage{amssymb}
\usepackage{amsmath}
\usepackage{amsthm}
\usepackage{booktabs}
\usepackage{enumitem}
\usepackage{graphicx}
\usepackage{color}
\usepackage{lscape}
\usepackage{tcolorbox}
\usepackage{longtable}

\newtcolorbox{promptbox}[1][]{colframe=black, colback=white, boxrule=0.5pt, arc=0pt, width=\linewidth, before skip=10pt, after skip=10pt}
\newcounter{promptcount}

\newcommand{\BibTeX}{B\kern-.05em{\sc i\kern-.025em b}\kern-.08em\TeX}

\begin{document}


\begin{frontmatter}


\paperid{XXX}


\title{A Comparative User Evaluation of XRL Explanations using Goal Identification}


\author[A]{\fnms{Mark}~\snm{Towers}\orcid{0000-0002-2609-2041}\thanks{Corresponding Author. Email: mt5g17@soton.ac.uk}}
\author[B]{\fnms{Yali}~\snm{Du}\orcid{0000-0001-5683-2621}}
\author[A]{\fnms{Christopher}~\snm{Christopher}\orcid{0000-0003-0305-9246}}
\author[A]{\fnms{Timothy}~J.~\snm{Norman}\orcid{0000-0002-6387-4034}}

\address[A]{School of Electronics and Computer Science, University of Southampton, UK}
\address[B]{Department of Informatics, Kings College London, UK}


\begin{abstract}
Debugging is a core application of explainable reinforcement learning (XRL) algorithms; however, limited comparative evaluations have been conducted to understand their relative performance. We propose a novel evaluation methodology to test whether users can identify an agent's goal from an explanation of its decision-making. Utilising the Atari's Ms. Pacman environment and four XRL algorithms, we find that only one achieved greater than random accuracy for the tested goals and that users were generally overconfident in their selections. Further, we find that users' self-reported ease of identification and understanding for every explanation did not correlate with their accuracy. 
\end{abstract}

\end{frontmatter}


\section{Introduction}
\label{sec:intro}
What should explanations achieve? Puiutta \emph{et al.} \cite{puiutta2020explainable} defines explainability as the fusion of two ideas: an accurate description of a model's inner workings, presented in a way that users can understand. It follows, therefore, that user surveys are a crucial component of XAI research for checking whether explanations are effective. One approach to measuring this is the ability for users to act upon an explanation or utilise it to answer a question, referred to as actionability \cite{gyevnar2025objective}. 

However, evaluating actionability in XRL presents unique challenges compared to explainable supervised learning due to the lack of labelled (correct) input/output pairs. As a result, prior research has primarily focused on evaluating whether users can predict an agent's next action through an explanation of its decision-making \citep{madumal2020explainable, silva2019optimization, alabdulkarim2210experiential, iyer2018transparency}. While understanding an agent's decision-making for their next action is important, RL agents are not optimised to (directly) maximise their next action; instead, they are trained to maximise their cumulative rewards over an episode. 

\begin{figure}[ht]
    \centering
    \includegraphics[width=\linewidth]{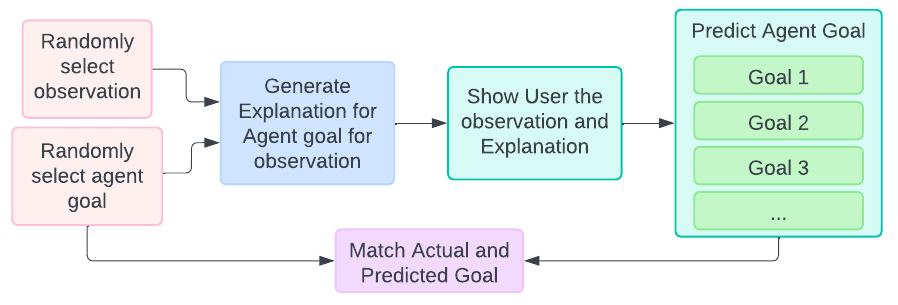}
    \vspace{-20px}
    \caption{Flowchart of the Goal Identification Task}
    \label{fig:goal-identification-flowchart}
\end{figure}

To address this evaluation gap, we propose a novel evaluation methodology centred on users predicting what goal an agent is pursuing, i.e., the reward function they are maximising over time. We refer to this as \textit{goal identification}. Our approach, illustrated in Figure \ref{fig:goal-identification-flowchart}, trains multiple agents in the same environment with unique reward functions, resulting in agents that complete different goals and use different decision-making. Then, selecting an agent at random with an explanation of its decision-making, users are asked to predict which goal the agent is completing (Section \ref{sec:survey-design}). This methodology simulates real-world debugging scenarios where practitioners need to understand: what (sub)goals agents are working towards, and why decisions were taken to achieve a given goal.

To demonstrate this methodology, we implement four distinct goals for the Atari Ms. Pacman environment \cite{bellemare2013arcade} and conduct a comprehensive user study with 100 participants. We evaluated four explanation mechanisms \cite{towers2024temporal, towers2025explaining, gyevnar2025objective, puri2020explain} across these goals, achieving 53.0\%, 34.9\%, 28.7\% and 22.5\% overall accuracy, respectively (Section \ref{subsec:user-accuracy}). We find that for two of the algorithms \cite{towers2025explaining, gyevnar2025objective}, the goal being explained affects their accuracy, with three goals being close to random and significantly greater for the other. Furthermore, we assess users' self-reported confidence in each goal prediction, finding a weak correlation to their accuracy and that users were often highly overconfident in their predictions (Section \ref{subsec:user-self-reported}). Then, for each algorithm, users rated their ease of identification and understanding using the explanations, for which we found no correlation with the user's accuracy. Overall, we demonstrate that self-reported metrics should not be relied on to evaluate an explanation's actionability and highlight that the tested explanations caused users to overestimate their goal-predictive abilities. A challenge that XRL must overcome to become applicable in the real world. 

In summary, this paper contributes the following to the state-of-the-art in XRL:
\begin{itemize}
    \item We propose a novel evaluation methodology for testing the explanation of an agent's decision-making if users can identify the agent's goal. We pair these objective questions with subjective questions to analyse whether algorithmic effectiveness correlates with user preference.
    \item Across 100 participants, we demonstrate that users' self-reported confidence, understanding and ease of identification are weakly correlated with their goal prediction accuracy and vary between explanation mechanisms. Further, we find that a user's average confidence is very weakly correlated with their accuracy.
\end{itemize}

\section{Comparative User Evaluation Design}
\label{sec:survey-design}
This section outlines our novel comparative user evaluation methodology, then its application in a user survey.

\subsection{Goal Identification}
\label{subsec:goal-id}
To debug an agent, it can be important to understand an agent's immediate influences; what state features cause agents to take what decision, or what state changes are necessary to achieve different behaviour. However, with RL's temporal nature, it is equally important to understand the context of how individual actions achieve an agent's overall objective/goal. As RL agents aim to maximise their rewards over time, an environment's reward function can be used to form a description of an agent's overall goal/objective. From this foundation, we propose that for a single environment, several distinct reward functions can be constructed to test whether explanations help users distinguish between them. This approach provides an objective ground truth (the actual goal used during training) while testing explanations on their long-term descriptive capability of an agent's decision-making.

For this methodology, the environment selected must support multiple meaningful and distinct reward goals or objectives. These goals can be encoded through the environment's reward function that agents learn to optimise with a separate agent trained for each. With a set of goals and associated agents, and a set of states, explanations of each agent's decision-making can be generated. This provides a set of states, goals, and explanations where a state-explanation pair can be selected, and users are evaluated if they can identify which goal the agent was pursuing (Figure \ref{fig:goal-identification-flowchart}). 


\subsection{Survey Structure and Design}
\begin{figure}[t]
    \centering
    \includegraphics[width=\linewidth]{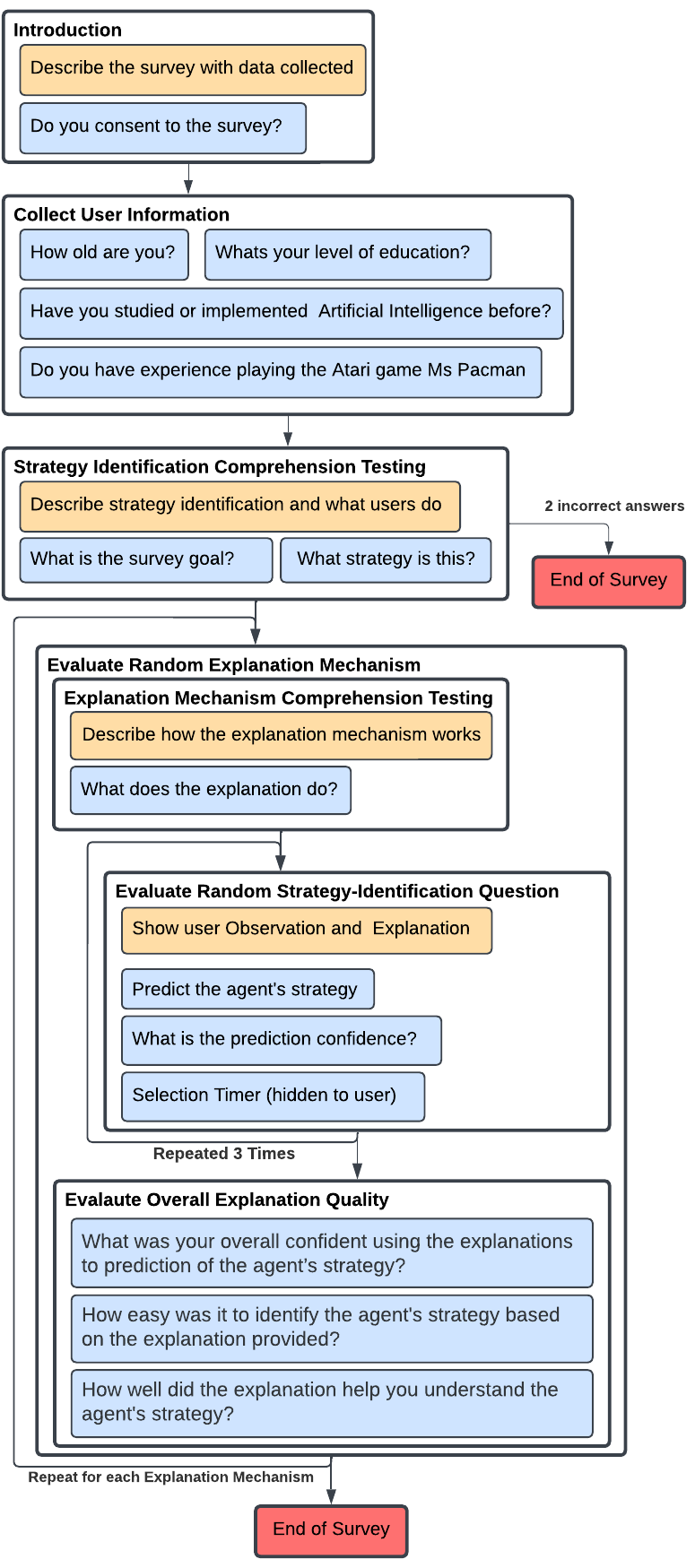}
    \vspace{-20px}
    \caption{Flowchart of the user survey. Blue boxes represent user questions, and the orange boxes represent information given to the user.}
    \label{fig:survey-structure}
\end{figure}

With the goal identification methodology outlined above, we incorporate it into our user survey as shown in Figure \ref{fig:survey-structure}. The first stage of the survey informs the user of the survey's purpose, user consent, and collects user information. Next, we conduct comprehension testing to ensure that users understand the survey, what goal identification is and what they have to do. Users who do not get at least one of the two questions correct are removed from the survey. 

For users who continue, we evaluate each explanation mechanism, first presenting an explanation summary and comprehension question to ensure that users understand the explanation. Next, users are shown four goal identification questions where, alongside selecting the agent's goal, we ask users to select their confidence in the prediction, from a 5-point Likert Scale \citep{likert1932technique} of ``Very Unconfident'' to ``Very Confident''. Additionally, we include a hidden timer to measure the time users take to answer each question. Once users have answered the four goal identification questions, the survey moves to elicit the user's opinions on the explanation mechanism. Again, we use a 5-point Likert Scale to measures users' overall confidence from ``Very Unconfident'' to ``Very Confident'', their ease of identifying the agent goals from ``Very Difficult'' to ``Very Easy'', and the user's general understanding of the explanation from ``Did Not Understand At All'' to ``Completely Understood''.

To control for possible influences on users, we include multiple comprehension gates, first in the introduction about the survey content and methodology, which can remove incorrect users, then for each explanation's content to ensure users understand what it explains. Additionally, we randomise the question and explanation presentation order per participant, and ensure that every question receives an equal number of evaluations, five, across the participant pool. 


\subsection{Survey Environment, Goals and Explanations}
\begin{figure}[ht]
    \centering
    \includegraphics[width=\linewidth]{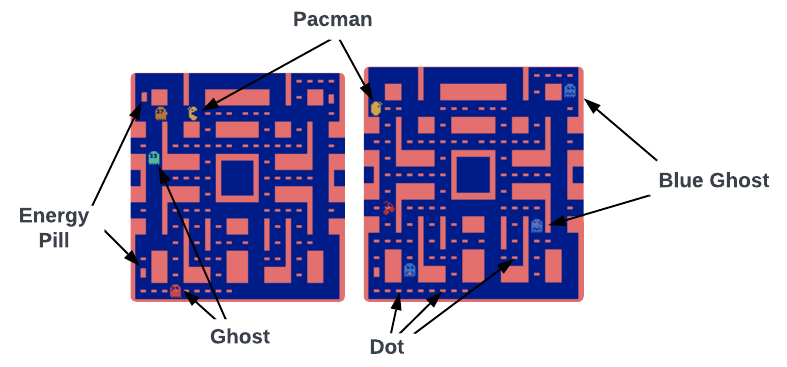}
    \vspace{-20px}
    \caption{Ms Pacman with the different reward sources labelled}
    \label{fig:mspacman-labelled}
\end{figure}

For the survey, we use Atari's Ms. Pacman environment \cite{bellemare2013arcade} as a well-known game that contains multiple reward sources, each labelled in Figure \ref{fig:mspacman-labelled}. We designed four goals with three criteria. First, strategic distinctiveness such that each goal should lead to observably different agent behaviours. Second, the goals should be easily understood by human participants and third, the goals would range from intuitive to counterintuitive. The goals implemented were:

\begin{itemize}
    \item \textbf{Eat Dots}: The agent receives +1 reward for each dot consumed. This encourages systematic dot collection while avoiding ghosts, representing the most intuitive Ms. Pacman strategy.
    \item \textbf{Eat Energy Pills and Ghosts (Eat EPaG}: The agent receives +1 reward for eating energy pills and +1 for each ghost consumed while they are blue. This creates a consume-then-hunt strategy.
    \item \textbf{Survive}: The agent receives +0.5 reward for each timestep survived (i.e., not losing a life). This encourages defensive play, maximising distance from ghosts and using energy pills strategically for protection.
    \item \textbf{Lose a Life}: The agent receives -0.5 reward for each timestep alive, incentivising rapid life loss. 
\end{itemize}

The first three goals represent rational strategies that humans might reasonably pursue when playing Ms. Pacman, while the fourth is an irrational behaviour from a human's perspective. We intend to test whether users are equally effective at spotting rational and irrational behaviour from a human's perspective, an important feature for real-world debugging.

For each goal, we train a standard DQN agent \cite{mnih2015human} for 10 million steps using an Impala vision encoder \cite{espeholt2018impala}, proven to improve performance \citep{clark2024beyond}. Videos demonstrating these agents, neural network weights, goal implementations, explanation mechanisms, and anonymised survey results are all available on Github\footnote{\url{https://github.com/pseudo-rnd-thoughts/eval-xrl-goal-identification}}.

We use four explanation algorithms for comparison that each explain different components of an agent's decision-making:
\begin{itemize}
    \item Dataset Similarity Explanation (DSE, \cite{towers2024temporal}) - A video of the agent's decisions from a similar observation within their training, presenting possible future observation-actions that the agent would take.
    \item TRD Summarisation (TRD Sum, \cite{towers2025explaining}[Section 6.1]) - A natural language description and figure of the agent's expected rewards for each of the next 40 timesteps explain what quantity and frequency of rewards the agent expects to receive.
    \item Specific and Relevant Feature Attribution (SARFA, \cite{puri2020explain}) - A saliency map image of the agent's decision-making that highlights observation components that most influence the agent's next action.
    \item Optimal Action Description (OAD, \cite{gyevnar2025objective}) - A natural language description of the agent's next action.
\end{itemize}

We selected these to provide two explanation mechanisms \cite{towers2024temporal} and \cite{towers2025explaining} that work to explain the future, and two explanation mechanisms that only explain an agent's decision-making in terms of their next action (saliency map and natural language description). Further, the explanations utilise a range of explanatory mediums: video, images and text.  

For the goal-identification questions, we generate 20 observations (five from each goal) with an explanation of each observation and goal producing 80 unique observation-goal questions. 

\section{Analysing Survey Results}
\label{sec:analyse-user-survey}
We conducted the user evaluation with 100 participants, using Prolific, a crowdsourcing website that provides access to participants worldwide, and Qualtrics for hosting and implementing our online surveys.\footnote{We obtained ethical approval for the survey from our institutional review board (ERGO FEPS/99644).} The pool of participants was filtered to those in North America and the UK whose first language is English and who have at least a High School GED / A-level. Each user was paid \pounds 5 to complete the survey, with the median completion time being 14 minutes and 8 seconds. The raw anonymised survey results are provided in the associated GitHub project \footnotemark[1]. From these results, we investigate various research questions regarding how well users identify an agent's goal given an explanation (Section \ref{subsec:user-accuracy}) and the subjective self-reported answers (Section \ref{subsec:user-self-reported}).

\subsection{Can Users Accurately Predict Agent Goals?}
\label{subsec:user-accuracy}
The evaluation methodology proposed in Section \ref{sec:survey-design} is centred on goal identification, where users select the goal description that most closely matches an explanation's description of an agent's decision-making. Table \ref{tab:mechanism-accuracy} lists the accuracy of each explanation mechanism across all agent goals. 

\begin{table}[ht]
    \centering
    \caption{Each explanation mechanism's accuracy across all agent goals.}
    \label{tab:mechanism-accuracy}
    \begin{tabular}{p{5cm}|p{2.5cm}}
        \textbf{Explanation mechanism} & \textbf{Average Participant Accuracy} \\ \hline
        Dataset Similarity Explanation & 53.0\% \\
        TRD Summarisation & 34.9\% \\
        Optimal Action Description & 28.7\% \\
        Specific and Relevant Feature Attribution & 22.5\% \\
    \end{tabular}
\end{table}

We find that the two explanation mechanisms that were designed to explain an agent's future outcomes (DSE and TRD Summarisation) had the highest accuracy of 53.0\% and 34.9\%, respectively. While OAD and SARFA, which primarily focused on the agent's next action, had the lowest accuracies of 28.7\% and 22.5\%, respectively. For users, guessing randomly would be expected to have 25\% accuracy. 

\begin{figure}[th]
    \centering
    \includegraphics[width=\linewidth]{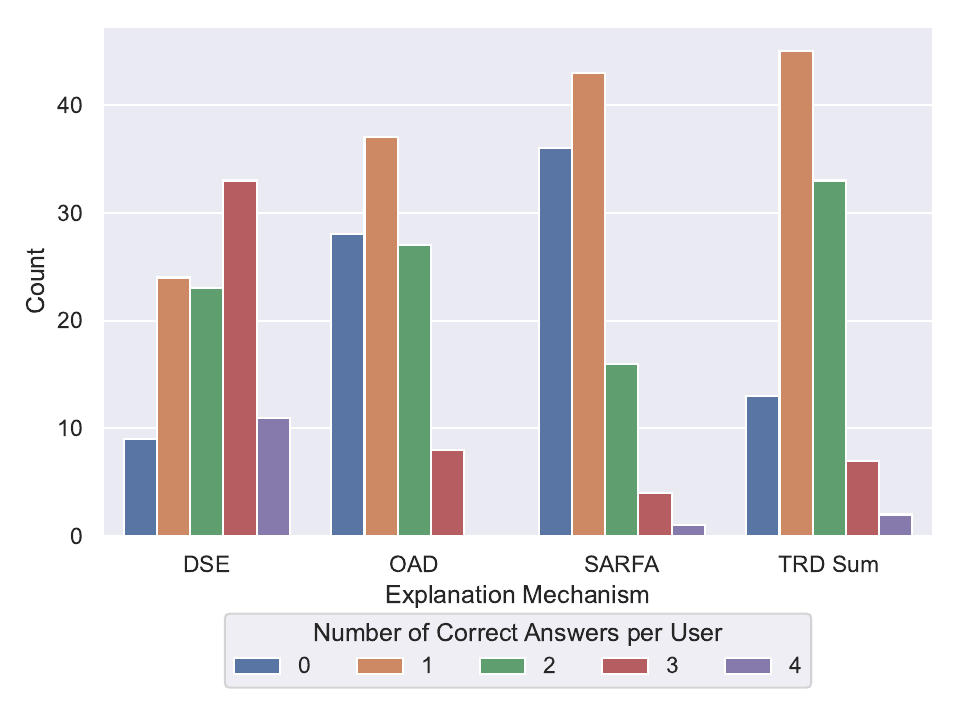}
    \vspace{-20px}
    \caption{The number of users who got each possible number of correct answers for each explanation mechanism.}
    \label{fig:exp-mech-user-accuracy}
\end{figure}

Investigating each user's performance for each explanation mechanism, we found the number of users who got 0, 1, 2, 3, or 4 answers correct for each explanation mechanism. Like the average accuracy, DSE and TRD Summarisation had the greatest number of users with at least one answer correct, at 91 and 87, respectively, compared to 72 and 67 for OAD and SARFA. However, DSE is the only algorithm where a consistent number of users could correctly identify 3 or 4 agent goals with 33 and 11, respectively, compared to 8, 5, and 9 for OAD, SARFA and TRD Summarisation, respectively, which got \emph{either} 3 or 4 correct answers.

\begin{figure}[th]
    \centering
    \includegraphics[width=\linewidth]{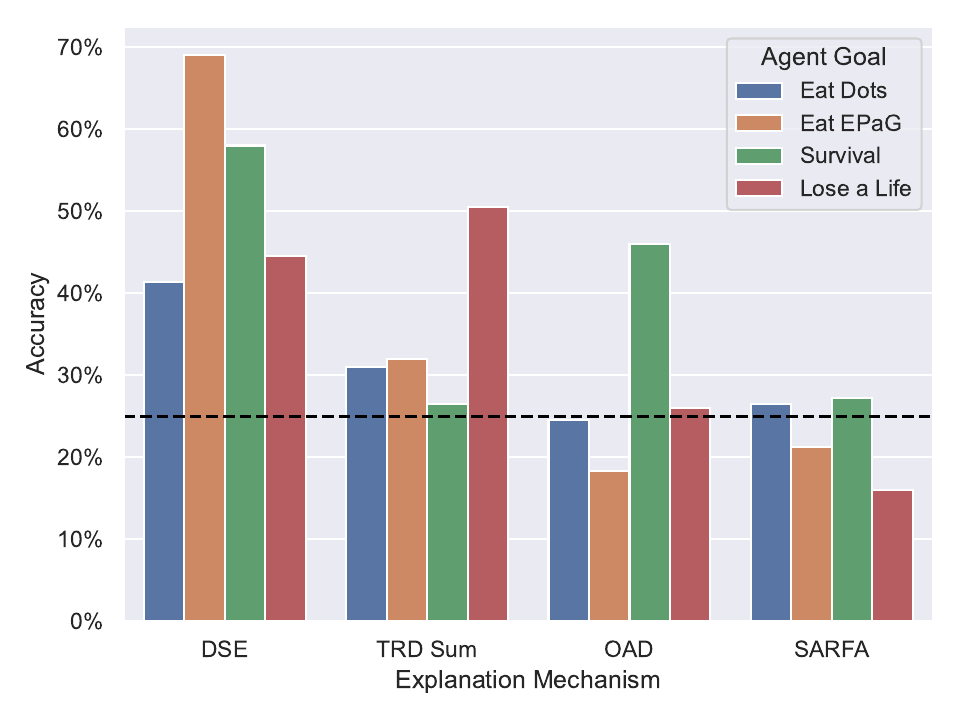}
    \vspace{-20px}
    \caption{A bar chart for each explanation mechanism with the accuracy of the explanation mechanism across each agent goal. Each bar's accuracy is averaged across 100 answers from users. The dotted line is 25\%, the expected accuracy for random guesses.}
    \label{fig:exp-mech-goal-accuracy}
\end{figure}

Is the accuracy of an explanation mechanism influenced by the agent's goal being explained? Figure \ref{fig:exp-mech-goal-accuracy} plots each explanation mechanism's accuracy for each goal. Ideally, all explanation mechanisms would achieve the same performance for every goal. However, we don't find that for any of the tested explanation mechanisms. For DSE, there is nearly a 30\% difference between goals with 69.3\% for Eat Energy Pills and Ghosts (Eat EPaG) and 40.5\% for Eat Dots. 

\begin{figure}[th]
    \centering
    \includegraphics[width=\linewidth]{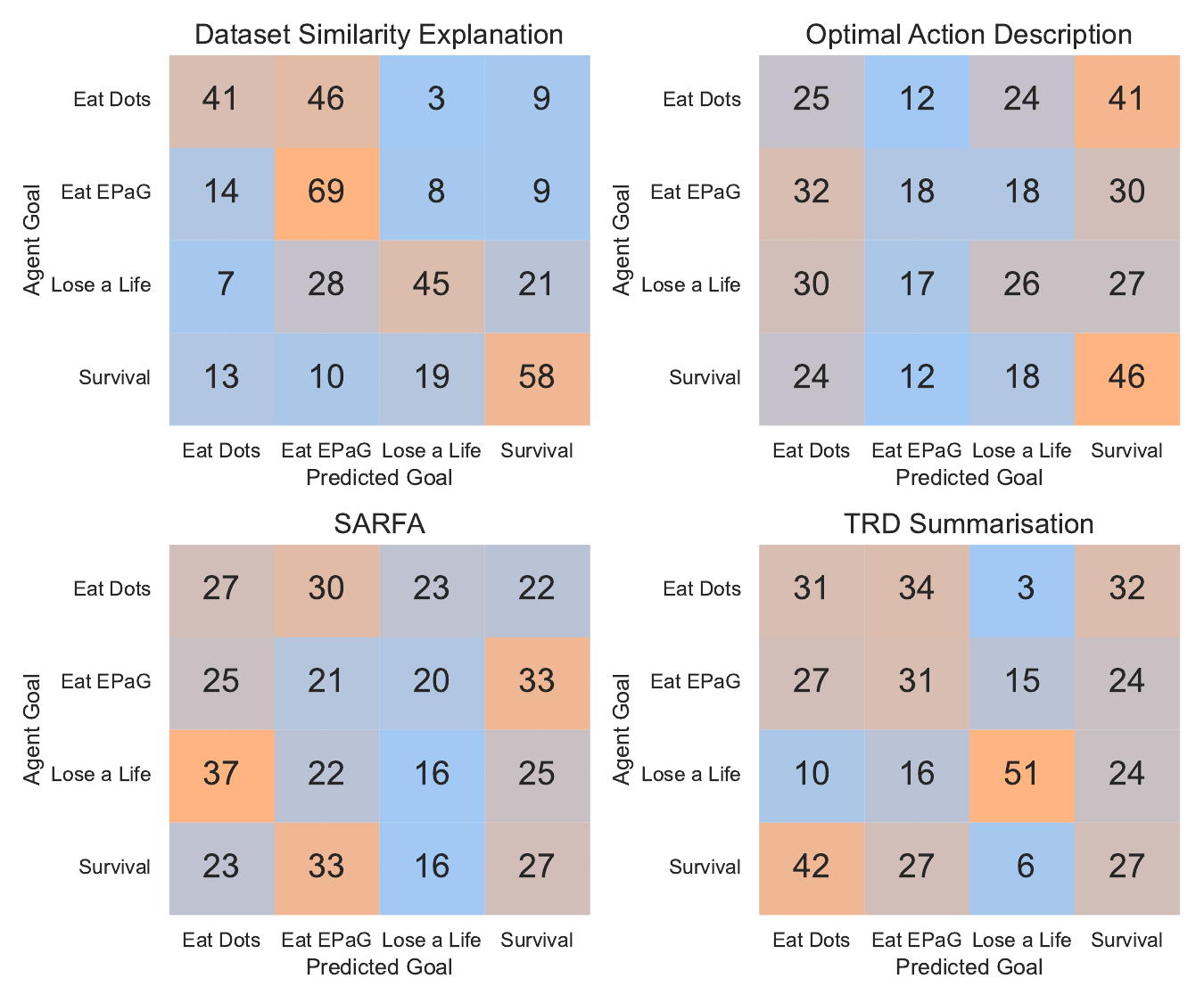}
    \vspace{-20px}
    \caption{Confusion matrix for the number of times the predicted and true agent goal for each explanation mechanism occurs, where the rows are the actual agent goal and the column is the predicted. Correct answers are found on the top left to bottom right diagonal. Each row will sum to 100 as the 20 observations are each evaluated five times.}
    \label{fig:exp-mech-goal-accuracy-confusion-matrix}
\end{figure}

Why the difference in performance? Plotting what users predicted for each goal, we can spot patterns (Figure \ref{fig:exp-mech-goal-accuracy-confusion-matrix}). For DSE, with the Eat Dots goal, users are highly likely to misidentify it as Eat EPaG (41 to 46 selections), while for the Lose a Life goal, users mistook the explanation for the Eat EPaG or Survival goal, with 28 and 21 selections compared to 45 for the correct answer. Overall, DSE struggle most for goals that have some overlap in strategy.

For TRD Summarisation and OAD, we find a significant difference between each agent's overall and individual goal accuracies. For three of the goals, their accuracy is close to random, while for Lose a Life and Survive, respectively, their accuracy increases to 50.5\% and 46.5\%. For TRD Summarisation, the reason for this difference in performance can be explained by the reward functions of the four goals. Eat Dots, Eat EPaG, and Survival all have positive rewards of varying quantities and frequencies, while the Lose a Life goal is the only reward function with negative rewards. This means that users could accurately predict the Lose a Life goal solely by observing the sign of future rewards. In comparison, the other three goals required additional attention and knowledge of the goal's reward functions to link the future rewards to a goal. 



For the Optimal Action Description (OAD), we find that users' accuracy was significantly higher in predicting the Survival goal compared to the other three goals; however, the reason for this is less obvious compared to TRD Summarisation. Figure \ref{fig:exp-mech-goal-accuracy-confusion-matrix} shows that users unevenly predict the Survival and Eat Dots goals with 144 and 111 total selections compared to 59 and 86 for the Eat EPaG and Lose a Life goals. We hypothesise that users had a sampling bias that they preferred to believe that agents were acting to Survive or Eat Dots rather than act to lose their lives, however, further research is required.

For SARFA \cite{puri2020explain}, we observe the lowest average accuracy of 22.5\%. A unique feature of user predictions for SARFA is that users consistently misinterpret the agent's decision-making for a different goal. From Figure \ref{fig:exp-mech-goal-accuracy-confusion-matrix}, we observe that for every goal, the user's most selected option is not the correct one, i.e., the Eat Dots goal is predicted as the Eat EPaG goal, Eat EPaG as Survival, Lose a Life as Eat Dots, and Survival as Eat EPaG. Why did users consistently mistake the agent's neural network's feature importance with different goals? For saliency map algorithms to be effective, humans and the neural networks must agree on what features are important for the same outcome, goals in this case. However, if there is a mismatch between the believed important features for an outcome, then users will consistently pick different goals from the actual goal. We observe this most strongly for the Lose a Life goal, where the most selected goal is the Eat Dots goal, with 37 compared to the next highest of 25.

\begin{figure}[ht]
    \centering
    \includegraphics[width=\linewidth]{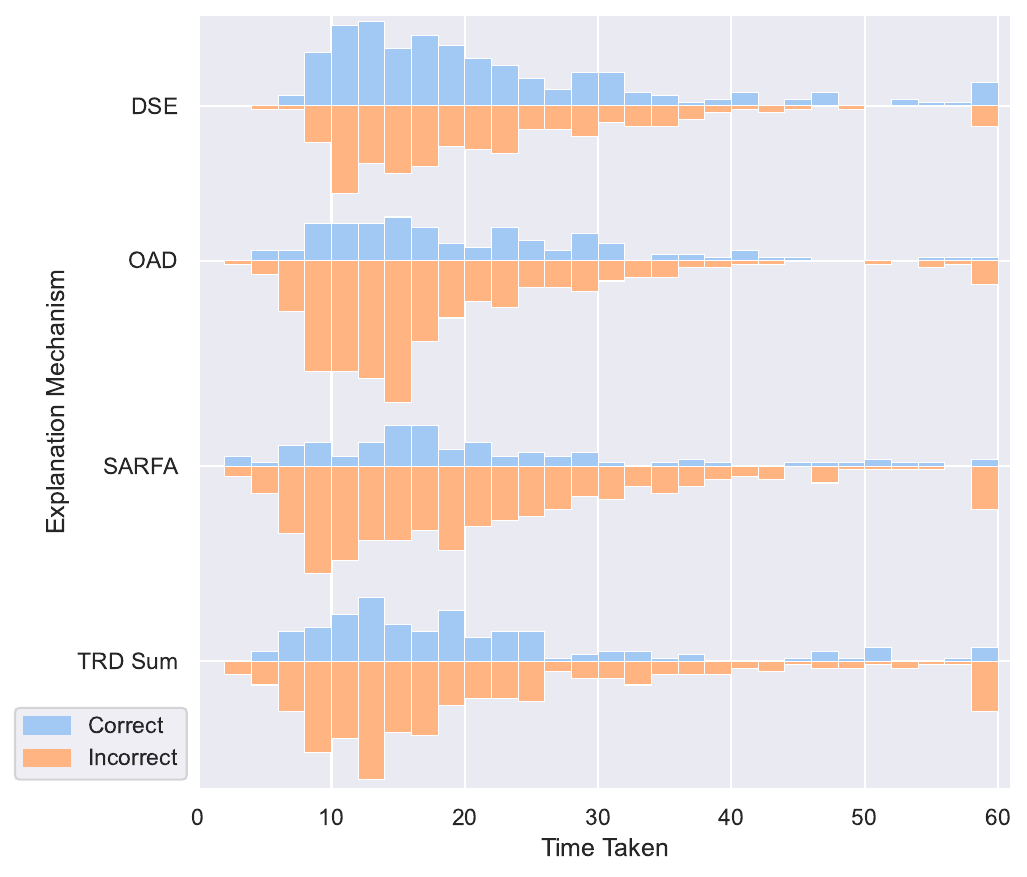}
    \vspace{-20px}
    \caption{A histogram of the time taken for each explanation mechanism. The time taken is clipped at 60 seconds.}
    \label{fig:exp-mech-time}
\end{figure}

\begin{table}[ht]
    \centering
    \caption{Table of p-values for each explanation mechanism of the time taken for correct and incorrect answers.}
    \label{tab:time-taken-accuracy-p-value}
    \begin{tabular}{p{2cm}|p{2.5cm}|c|c}
        \textbf{Explanation Mechanism} & \textbf{Number of correct / incorrect samples} & \textbf{Test Statistic} & \textbf{p-value} \\ \hline
        DSE & 213 / 187 & 0.055 & 0.900 \\
        OAD & 115 / 285 & 0.158 & 0.029 \\
        SARFA & 91 / 309 & 0.086 & 0.632 \\
        TRD Sum & 140 / 260 & 0.064 & 0.819 \\
    \end{tabular}
\end{table}

From Figures \ref{fig:exp-mech-goal-accuracy} and \ref{fig:exp-mech-goal-accuracy-confusion-matrix}, we have investigated each explanation mechanism's accuracy, but is this accuracy affected by the time users take to answer? Figure \ref{fig:exp-mech-time} plots a histogram of the time taken for each question split by whether the user was correct or not. To formally test this, we compute the p-value in Table \ref{tab:time-taken-accuracy-p-value} using the KS test \citep{massey1951kolmogorov} where the null hypothesis is that the time taken for correct and incorrect answers is from the same distribution. We set the significance level at 0.05 (5\%). 

\begin{figure}[th]
    \centering
    \includegraphics[width=\linewidth]{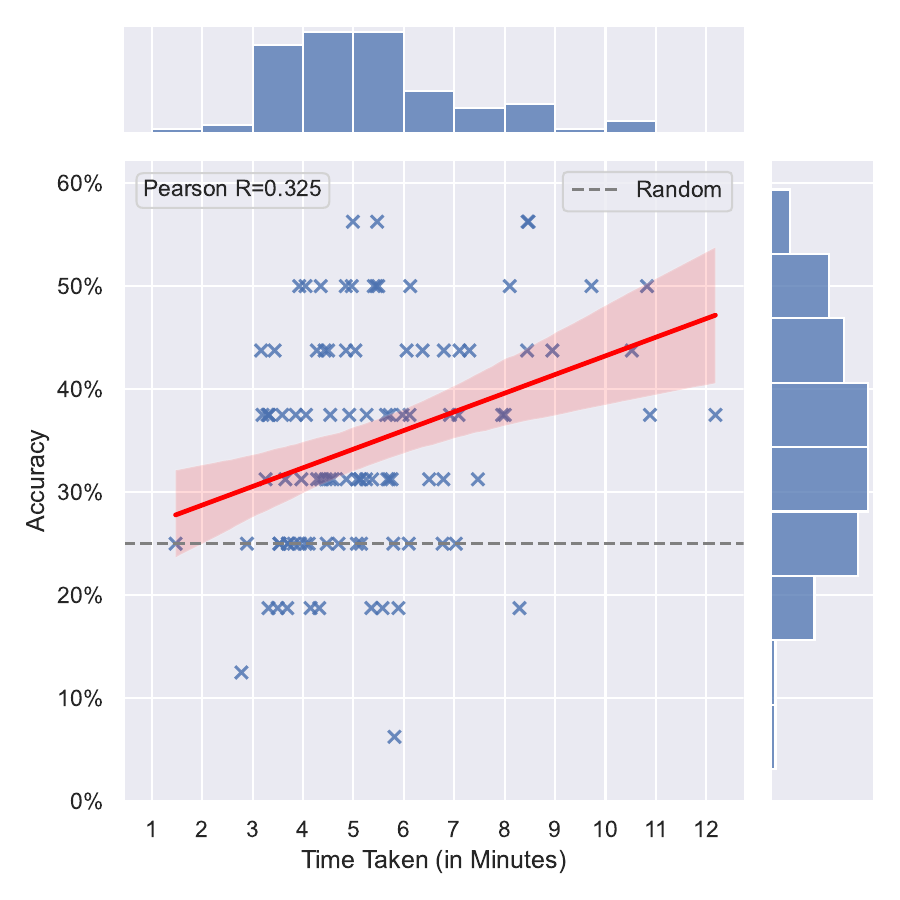}
    \vspace{-20px}
    \caption{A scatter plot of each user's cumulative time taken against their accuracy across all questions, with the red line being a plot of the linear regression. The individual question time taken is clipped at 60 seconds. The dashed line is at 25\%, representing the expected accuracy if users are selected randomly.}
    \label{fig:user-accuracy-time}
\end{figure}

For DSE, SARFA and TRD Summarisation, their p-values are greater than the significance level at 0.900, 0.632 and 0.819, respectively; thus, most likely, the time taken for correct and incorrect answers has the same distribution. However, for Optimal Action Description, the p-value is 0.029, meaning we reject the null hypothesis, and we cannot conclude that the time taken for incorrect and correct answers is the same distribution. Reviewing the cumulative density of OAD's answers, we find that the time taken by users was longer for correct than incorrect answers. This additional thinking time was most likely used to imagine how the agent's next action could accomplish each goal, and such that when users thought more deeply about the possible reasons for an action were, on average, more likely to be correct. Further, we tested whether the total time users took correlated with their overall accuracy, finding a Pearson R value of 0.325, indicating only a weak correlation (Figure \ref{fig:user-accuracy-time}). This means that for users who thought longer on every goal identification question did not have a significantly higher accuracy from those who selected answers more quickly. 

Finally, investigating the anonymised user characteristics collected from the survey (prior knowledge of AI or Ms. Pacman, age, gender, and education level), we trained a linear regression model to predict a user's accuracy from their reported characteristics. The coefficient of determination was $R^2=0.221$, where 1.0 equals the perfect prediction, meaning that no user characteristics was a predictor of their performance. We further investigated the Variance Inflation Factor \citep[Page 108]{james2013introduction} to measure multi-column correlations, similarly finding that no characteristic had a multicollinearity value greater than five, a standard cutoff value.

\subsection{What do Users believe about the Explanations?}
\label{subsec:user-self-reported}
Within the survey, alongside the goal identification questions, we ask users to self-report their confidence for each answer and, for each explanation mechanism, their overall confidence, ease of identification, and understanding. In this section, we first analyse this subjective data for each ``Which goal?'' question to understand the relationship between the user's confidence and accuracy. We then investigate how users' overall ratings of each explanation mechanism correlate with their accuracy. 

\begin{table*}[ht]
    \centering
    \caption{Table of Explanation Mechanism and user confidence with the user accuracy and their count in brackets.}
    \label{tab:strategy-confidence-accuracy}
    \begin{tabular}{p{3cm}|c|c|c|c|c}
        \textbf{Explanation Mechanism} & \textbf{Very Unconfident} & \textbf{Unconfident} & \textbf{Neutral} & \textbf{Confident} & \textbf{Very Confident} \\ \hline
        DSE & 80.0\% (5) & 31.2\% (32) & 48.9\% (92) & 55.9\% (213) & 60.3\% (58) \\
        OAD & 55.6\% (9) & 28.6\% (42) & 28.7\% (129) & 29.7\% (172) & 20.8\% (48) \\
        SARFA & 21.4\% (14) & 16.9\% (83) & 25.2\% (115) & 23.8\% (143) & 24.4\% (45) \\
        TRD Summarisation & 56.2\% (16) & 30.6\% (72) & 32.6\% (132) & 36.9\% (149) & 35.5\% (31) \\
    \end{tabular}
\end{table*}

\begin{table*}[ht]
    \centering
    \caption{Average user rating for each explanation mechanism on the three overall rating questions with the Likert Ratings discretised (1.0 for the lowest rating and 5.0 for the highest rating).}
    \label{tab:mech-overall-ratings}
    \begin{tabular}{p{3.5cm}|c|c|c}
        \textbf{Explanation Mechanism} & \textbf{Selection Confidence} & \textbf{Ease of Identification} & \textbf{Explanation Understanding} \\ \hline
        DSE & 3.48 & 3.23 & 3.56 \\
        OAD & 3.25 & 3.06 & 3.50 \\
        SARFA & 2.98 & 2.72 & 3.23 \\
        TRD Summarisation & 2.94 & 2.61 & 3.01 \\
    \end{tabular}
\end{table*}

\begin{figure}
    \centering
    \includegraphics[width=\linewidth]{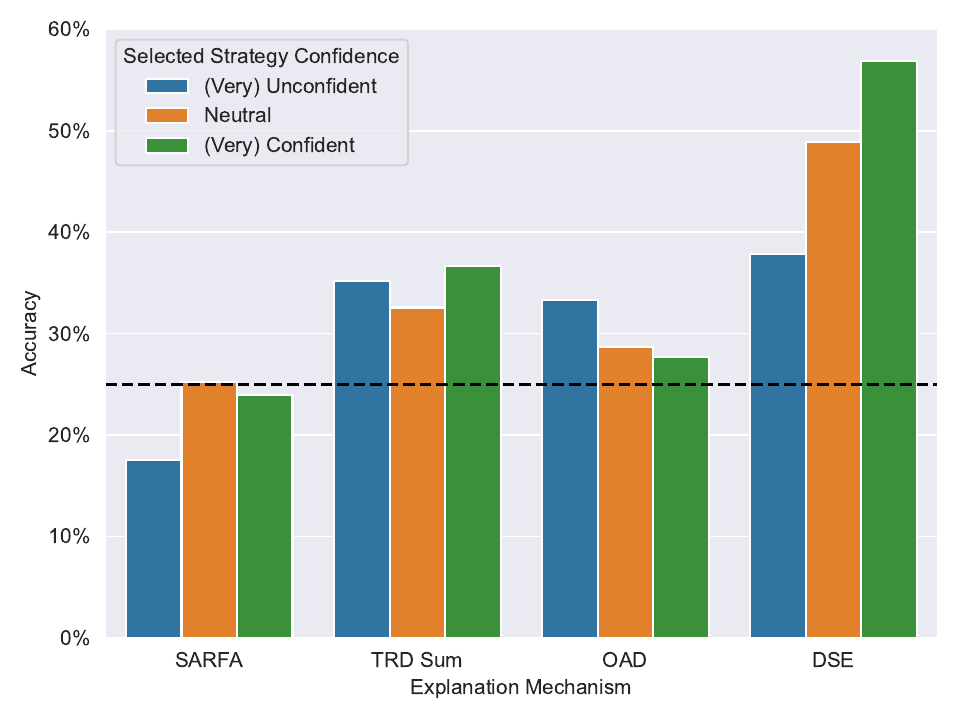}
    \vspace{-20px}
    \caption{Barchart of user accuracy with their selected confidence. Very unconfident and unconfident are grouped as (very) unconfident, and very confident and confident are grouped as (very) confident. The dashed line represents the expected accuracy with random guesses.}
    \label{fig:strategy-confidence-accuracy}
\end{figure}

For each goal identification question, users select the goal that matches the explanation content and their confidence using a 5-point Likert Rating between Very Unconfident and Very Confident. Table \ref{tab:strategy-confidence-accuracy} lists each explanation mechanism's accuracy for each confidence level. For Figure \ref{fig:strategy-confidence-accuracy}, to visualise Table \ref{tab:strategy-confidence-accuracy}, we group Very Confident and Confident into (Very) Confident due to the limited number of answers for Very Confident. We repeat this for Very Unconfident due to the same reason. 

For DSE, there is a clear correlation between the user's confidence and their accuracy, with each confidence rank increasing the expected accuracy. In contrast, OAD has the opposite confidence-accuracy correlation, with a decreasing accuracy for each increased confidence rank. For TRD Summarisation, user accuracy isn't correlated with accuracy, with the lowest number of users who selected Very Confident at 31 (out of 400) answers and the most at Neutral confidence, 136. For SARFA, it had the most users who were (Very) Unconfident at 97 (out of 400), for which they did have significantly lower accuracy at 17.5\% than the other confidence ranks, Neutral and (Very) Confident, at 25.2\% and 23.9\% accuracy respectively. However, this accuracy is still close to that of random selection at 25\%. 



\begin{figure}[ht]
    \centering
    \includegraphics[width=\linewidth]{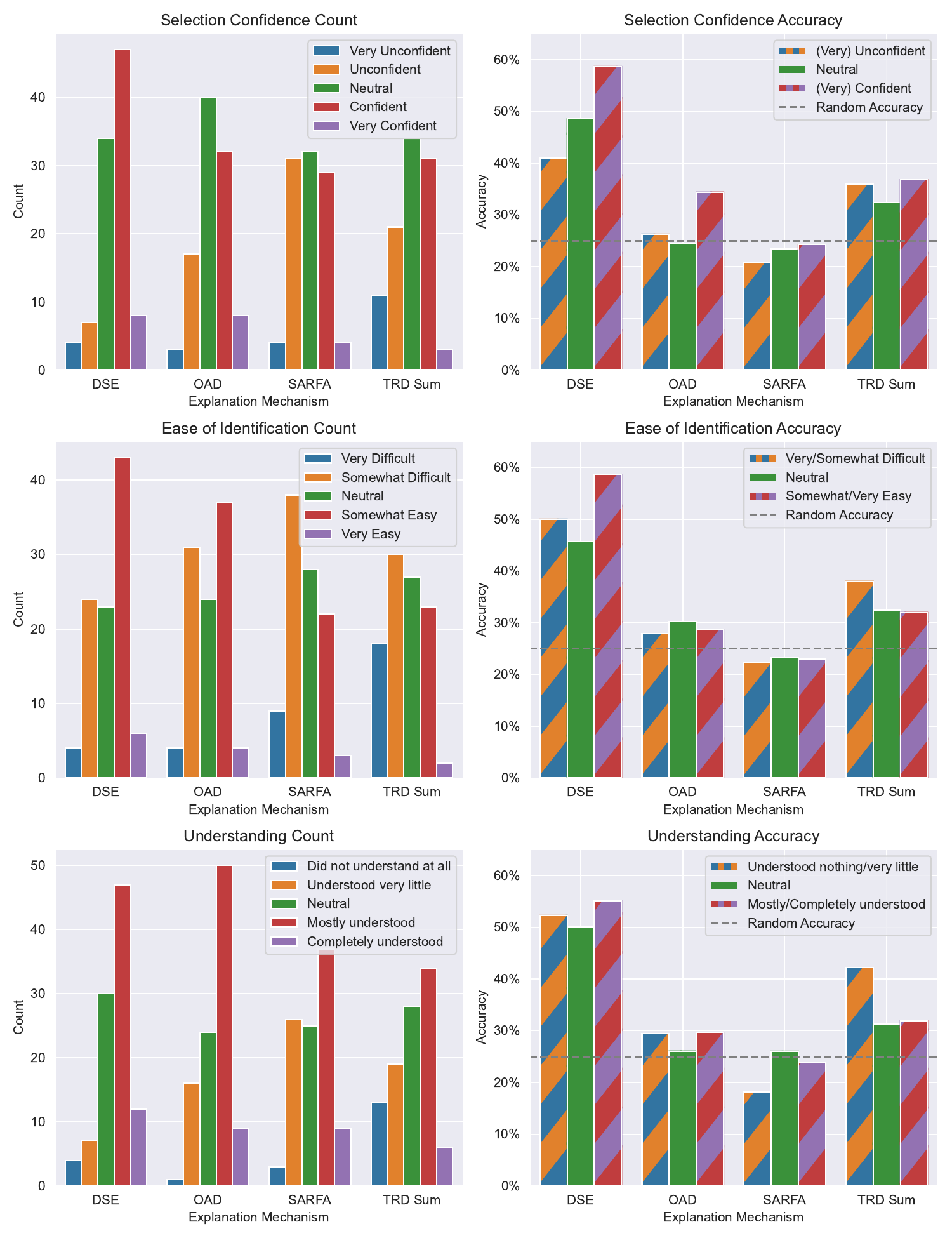}
    \vspace{-20px}
    \caption{Histogram of the number of times on the left and the accuracy for each explanation mechanism's confidence, ease of identification, and understanding is selected by users. }
    \label{fig:mech-overall-ratings}
\end{figure}

After answering each explanation mechanism's goal identification questions, users select their overall confidence, ease of identification or understanding of the explanations. We investigate how users rate each explanation mechanism and how it correlates with their accuracy. Figure \ref{fig:mech-overall-ratings} plots the number of users who select a Likert Rating and the number of questions correctly answered.\footnote{Due to the limited numbers of users who rated any explanation mechanism as either the highest or lowest Likert Rating, e.g., Very Confident or Very Unconfident, Figure \ref{fig:mech-overall-ratings} combines these selections into the second highest and lowest ratings.} For the overall explanation confidence, we find very similar results to the individual goal confidence ranks for each explanation. Viewing the number of selections of each confidence rank, ignoring the accuracy, we can see that SARFA and TRD Summarisation had a significantly greater number of users who found they were Very Unconfident and Unconfident, while DSE has the highest number of people who were Very Confident. 


Compared to the goal identification questions, in the overall ratings for each explanation mechanism, we asked two more questions: ``How easy was it to identify the agent's goal based on the explanation provided?'' and ``How well did the explanation help you understand the agent's goal?''. Like the overall user confidence, TRD Summarisation had significantly greater numbers of users who selected that they found the explanation ``Very Difficult'' in ease of identification for the goal and ``Did not understand anything at all'' from the explanation. This explanation is the most technical, requiring a basic level of understanding about what a reward means and how agents learn to maximise their rewards over time, possibly explaining why users found the explanation so difficult. For DSE and OAD, we find similar user preferences for mechanisms regarding ease of identification and understanding across all ranks. Despite these similarities in self-reported user beliefs, we, again, find that these beliefs do not correlate with the average user's accuracy, with DSE and OAD having a +20\% performance difference for all self-reported ranks.

\section{Related Work}
\label{sec:related-work}
Surveying 74 XRL papers, Milani \emph{et al.} \cite{milani2024explainable} found that only 16 included user studies with a wide variety of evaluation approaches being taken. In response to this, Gyevnar and Towers \cite{gyevnar2025objective} outline a range of evaluation approaches for XLR focusing on those with objective measures. One of the proposals they discuss is based on the evaluation approach we outline in Section \ref{subsec:goal-id}; however, they don't propose a survey structure or conduct a survey. 

Of the user surveys conducted, we are aware of two that evaluate explanation capabilities to explain an agent's long-term beliefs. 
Septon \emph{et al.} \cite{septon2023integrating} proposed a comprehensive evaluation methodology that systematically compares the effectiveness of combining local and global explanations for reinforcement learning agents. Their approach centres on designing an ordered preference list of reward functions, ranging from optimal human behaviour to irrational. Then, taking a random pair of agents trained on the reward functions, users have a binary choice ot select the agent they believe is better. 

Huber \emph{et al.} \cite{huber2023ganterfactual} introduced GANterfactual-RL, a novel approach for generating visual counterfactual explanations using generative adversarial networks to illustrate what minimal visual changes would cause an agent to choose different actions. Their evaluation framework focuses on users' ability to distinguish the most important reward sources (e.g., dots, energy pills, ghosts) between agents trained on different weighting of the reward sources. Users were asked to select the top two most important reward sources, corresponding to their weighting, from a list of options. 

Like our goal identification methodology, both rely on agents that were trained using different reward functions, but differ on the question asked to users: ``what goal is the agent pursuing?'', ``what agent's strategy do you prefer?'' \cite{septon2023integrating}, and ``what reward (sources) does the agent prioritise?'' \cite{huber2023ganterfactual}. Additionally, all three methodologies converge on concerning findings: users consistently demonstrate poor objective performance despite high subjective confidence, and self-reported measures show weak or no correlation with actual task accuracy. This convergence across different explanation types, task designs, and environments suggests a systematic challenge in XRL evaluation that transcends specific methodological choices. 

\section{Discussion}
\label{sec:eval-survey-discussion}


Taking inspiration from the application of explainability to debug/understand a trained RL agent, either pre-deployment to assess for flaws or post-deployment to identify why a decision was taken. We devised a novel evaluation methodology, proposed in Section \ref{sec:survey-design}, which we conducted with 100 participants and analysed their answers in Section \ref{sec:analyse-user-survey}. 

Overall, the Dataset Similarity Explanation \cite{towers2024temporal} had the highest accuracy (53.0\%) and user ratings. Meanwhile, the TRD Summarisation explanation \cite{towers2025explaining} had the second highest accuracy of 34.9\% but the lowest overall user ratings for every category (Table \ref{tab:mech-overall-ratings}). Both algorithms were designed to explain an agent's long-term decision-making, while Optimal Action Description \cite{gyevnar2025objective} and SARFA \citep{puri2020explain} were designed to only explain an agent's next action; however achieve performance close to expected random performance at 28.7\% and 22.5\%. 

Overall, we observe the following features from the survey:
\begin{itemize}
    \item \textbf{Low accuracy}: The average user accuracy was 34.9\%, with the lowest being 6.3\% and the highest being 56.3\%. Against an expected random accuracy of 25.0\%, only one of the explanation mechanisms (DSE) achieves accuracy success significantly above random for all agent goals. While we have not tested a wide variety of explanation mechanisms, we believe this survey demonstrates that there are still substantial performance gaps within the literature. 
    \item \textbf{Self-reported questions cannot determine effectiveness}: Comparing users' answers for the self-reported and objective questions of ``what goal?'', we find a weak correlation between user confidence and accuracy, and no correlation for most of the overall explanation ratings and accuracy. These results highlight a graver issue: not just that self-reported measures can only specify an explanation preference, but that these answers provide very weak or even misleading information on the explanation's effectiveness. This poses further problems in analysing user survey results from the literature, as only a few (though growing) number of papers conduct user surveys with objective user questions.
    \item \textbf{Overconfidence}: Like self-reported questions, user confidence is a critical feature to correlate correctly with user accuracy. This possibly poses a critical problem for applying explanations in real-world situations where misunderstanding an explanation can result in significant consequences. Ideally, in XRL, we hope that users only select that they have high confidence in an answer when they are correct.
\end{itemize}


Finally, reflecting on the survey, if we conducted it again, we would consider changing the survey in the following ways:
\begin{itemize}
    \item \textbf{Improve survey/explanation descriptions}: In the optional thoughts from users, several noted that they were either confused by what they needed to do in the survey or by the explanations themselves. If conducted again, further time would have been spent optimising the descriptions of the explanation mechanisms to ensure that the users could interpret the explanation more effectively.
    \item \textbf{Increase the number of questions asked}: The median time for users was 14 minutes to complete the survey, including 16 goal identification questions across four explanation mechanisms. With only four questions per explanation mechanism, this provides relatively little data that is more sensitive to user mistakes. Therefore, if conducted again, we would explore increasing the number of goal identification questions for each explanation mechanism to five or six; alternatively, reducing the number of explanation mechanisms shown each user to three while increasing the number of questions per explanation to ten or twelve. This should make the results more resilient to user mistakes.
    \item \textbf{Add attention checks}: For this survey, we included comprehension questions at the beginning to check that users understood the survey. However, we did not include attention checks throughout the survey to validate that users consistently checked and read the prompts/explanations. Adding such checks could help prevent super-speedy users from completing the survey without the necessary attention.
    \item \textbf{Policy Optimality}: We know from observing policy rollouts (videos of the agents) that they occasionally take sub-optimal actions from a human's perspective with knowledge of the agent's goal. It is difficult to assess, but the sub-optimality of agents could cause issues explaining behaviour that is sub-optimal from a human's perspective. The use of more powerful training algorithms, e.g. Rainbow DQN \citep{hessel2018rainbow} or Beyond the Rainbow \citep{clark2024beyond} that are trained for longer, e.g., 200 million steps, might minimise this issue if it exists.
\end{itemize}



\begin{ack}
This work was supported by the UKRI Centre for Doctoral Training in Machine Intelligence for Nano-electronic Devices and Systems [EP/S024298/1] and RBC Wealth Management.
\end{ack}



\bibliography{references}

\end{document}